%% file: main.tex
\documentclass[10pt,twocolumn,letterpaper]{article}

\usepackage[pagenumbers]{iccv} 
\usepackage{multirow}
\usepackage{makecell}
\usepackage{tikz}
\usepackage{mathrsfs}
\usetikzlibrary{positioning}
\usepackage{amsthm}
\usepackage{float}
\usepackage[utf8]{inputenc}
\usepackage{graphicx}
\usepackage{caption}
\usepackage{subcaption}
\usepackage{diagbox}


\definecolor{iccvblue}{rgb}{0.21,0.49,0.74}
\usepackage[pagebackref,breaklinks,colorlinks,allcolors=iccvblue]{hyperref}

\def\paperID{*****} 
\def\confName{ICCV}
\def\confYear{2025}

\title{Empirical Evidences for the Effects of Feature Diversity in Open Set Recognition and Continual Learning}

\author{Jiawen Xu\\
TU Berlin\\
Ernst-Reuter-Platz 7, Berlin, Germany\\
{\tt\small jiawen.xu@campus.tu-berlin.de}
\and
Odej Kao\\
TU Berlin\\
Ernst-Reuter-Platz 7, Berlin, Germany\\
{\tt\small odej.kao@tu-berlin.de}
}

\begin{document}
\maketitle
\input{sec/0_abstract}

\input{sec/1_intro}

\input{sec/2_related}
\input{sec/controlled_experiments}
\input{sec/7_conclusion}

{
    \small
    \bibliographystyle{ieeenat_fullname}
    \bibliography{main}
}

\end{document}

%% file: sec/0_abstract.tex
\begin{abstract}

Open set recognition (OSR) and continual learning are two critical challenges in machine learning, focusing respectively on detecting novel classes at inference time and updating models to incorporate the new classes. While many recent approaches have addressed these problems—particularly OSR—by heuristically promoting feature diversity, few studies have directly examined the role that feature diversity plays in tackling them. In this work, we provide empirical evidence that enhancing feature diversity improves the recognition of open set samples. Moreover, increased feature diversity also facilitates both the retention of previously learned data and the integration of new data in continual learning. We hope our findings can inspire further research into both practical methods and theoretical understanding in these domains.
\end{abstract}

%% file: sec/1_intro.tex
\section{Introduction}
\label{sec-introduction}

Deep neural networks have exhibited remarkable prowess in classification tasks, spanning domains such as video classification ~\cite{karpathy2014large}, sentiment analysis ~\cite{zhang2018deep}, and fault diagnosis ~\cite{lei2020applications}. 
Nevertheless, practical applications often entail challenges wherein the samples of semantic categories may be difficult to exhaust or may dynamically evolve over time, diverging from the training data. To update the models for these novel data, instead of training from scratch, techniques such as continual learning have been proposed, which can greatly reduce the training time and memory requirements for the historical data \cite{kirkpatrick2017overcoming, zenke2017continual, rebuffi2017icarl, buzzega2020dark, serra2018overcoming, wang2022learning, li2024contrastive}. 
Besides, in order to enable the models to know not only how to update the models but also when to start the new training, the models are supposed to be able to detect the novel data during inference. Approaches for this are normally under the categories of open set recognition (OSR) ~\cite{hassen2020learning, dhamija2018reducing, chen_2020_ECCV}, out-of-distribution detection (ODD) \cite{yang2022openood, yang2024generalized, sun2022dice}, or novelty detection \cite{pimentel2014review, blanchard2010semi, chen2021novelty}. These two aspects should hence naturally work and be researched together \cite{xu2023openincrement}. In this study, we look into continual learning, and more specifically, class-incremental learning, and open set recognition. 

Apart from numerous novel approaches proposed in recent years for these two topics, some studies have started to investigate from the fundamental perspectives, such as the loss landscape \cite{fang2024revisiting, bian2024make, deng2009imagenet} and feature diversity \cite{li2022provable, wang2024exploring}. For feature diversity, \cite{wang2024exploring} provides theoretical proof and lacks of precise empirical observations. \cite{li2022provable} focuses on the setting of the frozen feature extractor trained in the initial task for the following continual learning tasks. 
In this work, we seek to provide empirical evidence of feature diversity for both open set recognition and continual learning. We conduct controlled experiments on the synthetic dataset, which can manipulate the models for the desired features to learn. We then analyze the models with the metrics for open set recognition, feature forgetting, and forward transfer. The results show that learning diverse features is beneficial for improving the models for detecting the open set samples, restraining features from the observed data, as well as learning new tasks.

Throughout this study, we use open-sets and close-sets to name open-set and close-set samples respectively. The observed data is for the data in the past tasks in continual learning. We focus on class-incremental learning only, hence the observed data are classes that are disjoint from the current data.

%% file: sec/2_related.tex
\section{Related Work}
\label{sec-related}
The most related work to this study is open set recognition and continual learning. We briefly introduce these two topics here.

\noindent\textbf{Open Set Recognition} OSR is concerned with identifying samples from novel classes during inference.
OSR methods for discriminative models can be broadly categorized into two main groups, namely synthesizing outliers ~\cite{asg_Yu17, GOpenMax_Ge17, neal2018open} and learning more discriminative models \cite{hassen2020learning, dhamija2018reducing, yang2020convolutional, perera2020generative, millerclass, chen_2020_ECCV}. 
Besides relying solely on discriminative classifiers, a bunch of works apply extra models, e.g., generative models, to learn representations of the in-sets \cite{oza2019c2ae, yoshihashi2019classification, zhang2020hybrid, cao2021open}. 

\noindent\textbf{Continual Learning} The main task of continual learning is to mitigate catastrophic forgetting \cite{mccloskey1989catastrophic}.
Conventional continual learning approaches can fall into three categories: regularization-based \cite{kirkpatrick2017overcoming, li2017learning, aljundi2018memory}, reply-based \cite{rebuffi2017icarl, hou2019learning, kang2022class}, and architecture-based \cite{li2019learn, hu2019overcoming, hung2019compacting}. 

%% file: sec/controlled_experiments.tex
\section{Experiments} \label{sec-exp}

\subsection{General Settings}  \label{subsec-general-settings}
\begin{figure}[htbp]
\centering
\centering
\fbox{\includegraphics[width=0.1\textwidth]{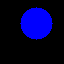}}
\fbox{\includegraphics[width=0.1\textwidth]{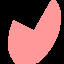}}
\fbox{\includegraphics[width=0.1\textwidth]{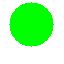}}
\fbox{\includegraphics[width=0.1\textwidth]{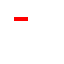}} \\[0.5em]
\fbox{\includegraphics[width=0.1\textwidth]{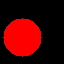}}
\fbox{\includegraphics[width=0.1\textwidth]{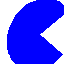}}
\fbox{\includegraphics[width=0.1\textwidth]{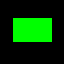}}
\fbox{\includegraphics[width=0.1\textwidth]{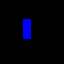}}
\captionof{figure}{Examples from the synthetic dataset. Each combination of the color and shape forms a class of data.
All backgrounds are set to be black or white.}
\label{fig-three-toy}
\end{figure}

In order to investigate the connection between feature diversity and continual learning as well as continual learning, we synthesized datasets for supervised classification, where the features learned by the models can be known (It will be introduced in the detailed experiments later). Our synthesis datasets consist of two types of features, i.e., colors (blue, red, green, and pink) and shapes (circle, rectangle, and ellipse). The position and size of the shapes are randomly generated.
Each combination of the color and shape forms a class of data.
To eliminate the background influence, the background of all images is set to either black or white.
It will be introduced in the following sections that the black background images are used for training, and the white background data is utilized for testing in some cases.
The spatial size of the images is $64*64$.
Examples from the datasets are shown in Figure \ref{fig-three-toy}.

Considering the simplicity of the synthesis datasets, we design and use a small-scale neural network for all experiments in this study. It is a five-layer CNN, and the network architecture configurations are in Table \ref{tab-model-architecture}. For all experiments in the following text, there are 100 and 50 images in each class in the training and testing sets, respectively.

\begin{table}[h]
\centering
\scriptsize
\begin{tabular}{ c|ccc } 
 \toprule
Layer ID & Type & Input & Output \\
 \midrule
0 & Conv2D & 64*64*3 & 64*64*10 \\ 
1 & AvgPooling2D & 64*64*10  &  32*32*10      \\
2 & Flatten   &  32*32*10 &    10240 \\
3 & Linear  &   10240  & 1000    \\
4 & Linear  &   1000  & 20    \\
5 & Linear  &   20  & num-classes    \\
 \bottomrule
\end{tabular}
\caption{List of the network architecture configurations, layer type, in-and output dimensions, that were applied in all experiments in this study.}
\label{tab-model-architecture}
\end{table}

\subsection{Open Set Recognition} \label{subsec-osr}

\begin{figure}[t]
\centering
  \begin{subfigure}{0.08\textwidth}
    \centering
    \includegraphics[width=.99\linewidth]{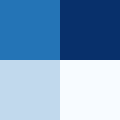}
    \caption{E1, 0}
    \label{fig-confusion-11}
  \end{subfigure}
  \vspace{0.2cm}
  \begin{subfigure}{0.08\textwidth}
    \centering
    \includegraphics[width=.99\linewidth]{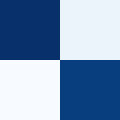}
    \caption{E1, 10}
    \label{fig-confusion-12}
  \end{subfigure}
  \vspace{0.2cm}
  \begin{subfigure}{0.08\textwidth}
    \centering
    \includegraphics[width=.99\linewidth]{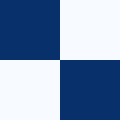}
    \caption{E1, 30}
    \label{fig-confusion-13}
  \end{subfigure}
  \vspace{0.2cm}
  \begin{subfigure}{0.08\textwidth}
    \centering
    \includegraphics[width=.99\linewidth]{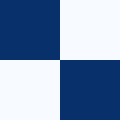}
    \caption{E1, 50}
    \label{fig-confusion-14}
  \end{subfigure}
  \vspace{0.2cm}
  \begin{subfigure}{0.08\textwidth}
    \centering
    \includegraphics[width=.99\linewidth]{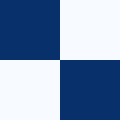}
    \caption{E1, 100}
    \label{fig-confusion-15}
  \end{subfigure}

  \medskip

  \begin{subfigure}{0.08\textwidth}
    \centering
    \includegraphics[width=.99\linewidth]{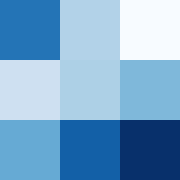}
    \caption{E2, 0}
    \label{fig-confusion-21}
  \end{subfigure}
  \vspace{0.2cm}
  \begin{subfigure}{0.08\textwidth}
    \centering
    \includegraphics[width=.99\linewidth]{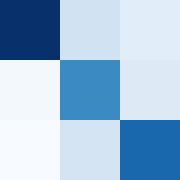}
    \caption{E2, 10}
    \label{fig-confusion-22}
  \end{subfigure}
  \vspace{0.2cm}
  \begin{subfigure}{0.08\textwidth}
    \centering
    \includegraphics[width=.99\linewidth]{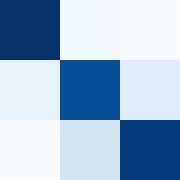}
    \caption{E2, 30}
    \label{fig-confusion-23}
  \end{subfigure}
  \vspace{0.2cm}
  \begin{subfigure}{0.08\textwidth}
    \centering
    \includegraphics[width=.99\linewidth]{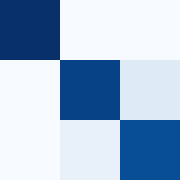}
    \caption{E2, 50}
    \label{fig-confusion-24}
  \end{subfigure}
  \vspace{0.2cm}
  \begin{subfigure}{0.08\textwidth}
    \centering
    \includegraphics[width=.99\linewidth]{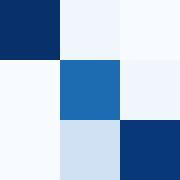}
    \caption{E2, 100}
    \label{fig-confusion-25}
  \end{subfigure}
  \caption{Confusion matrices during training in E1 (first row) and E2 (second row) of epoch 0, 10, 20, 30, 50, and 100 (deeper color indicates higher value). It can be found that, in E1, the two classes can be well discriminated after epoch 10. Whereas in E2, the model is easier to be confused by red rectangles and red circles, for which the shapes should be learned.}
  \label{fig-confusion}
\end{figure}

We train two supervised classifiers (E1 and E2) in this section. E1 is a binary classifier in which the two classes to discriminate are blue circles and red rectangles.
The class settings are listed in Table \ref{tab-setting1}.
It is straightforward to find that the model can perfectly distinguish between the two classes through the color only. And it is well known that CNNs exhibit a bias toward color over shape \cite{geirhos2018imagenet,singh2020assessing}. Thus, the model in E1 relies primarily on color features.
In E2, a third class of red circles is introduced. In this case, accurate classification of all three classes requires the model to incorporate shape information. Therefore, E2 models are encouraged to learn more comprehensive features than in E1. Both models are trained for 100 epochs. The testing accuracies of E1 and E2 are $100\%$ and $99.5\%$, respectively.
To better illustrate the feature reliance during the training progress, Figure \ref{fig-confusion} depicts the confusion matrices at different epochs. The two close-set classes in E1 can be quickly and completely well classified in a few epochs. However, the red rectangles and red circles (the third class) are always confused in E2 till the end of the early tens of epochs whereas the first two classes can be more easily to be classified as in E1. It verifies that E2 requires learning more features than E1, and the shapes are harder to learn than color, which aligns with \cite{geirhos2018imagenet}.

\begin{table}[ht]
  \centering
  \scriptsize
  \begin{subtable}[t]{0.2\textwidth}
    \centering
    \caption{OSR}
    \begin{tabular}{ccc}
    \toprule
    Index & Close-set &  Open-set     \\
    \midrule
    E1  & \begin{tikzpicture}[baseline=(current bounding box.center)]
   \draw[fill=blue!100,draw=blue] (-1.2,0) circle [radius=0.18];
  \draw[fill=red!100,draw=red] (-0.85,-0.18) rectangle ++(0.36,0.36);
\end{tikzpicture} & \begin{tikzpicture}[baseline=(current bounding box.center)]
  \draw[fill=blue!100,draw=blue] (-0.7,-0.18) rectangle ++(0.36,0.36);
  \draw[fill=green!100,draw=green] (-0.2,-0.18) rectangle ++(0.36,0.36);
  \draw[fill=green!100,green=blue] (-0.6,-0.5) circle [radius=0.18];
  \draw[fill=blue!100,draw=blue] (0,-0.5) ellipse [x radius=0.29, y radius=0.2];
  \draw[fill=pink!100,draw=pink] (-0.3,-0.9) ellipse [x radius=0.29, y radius=0.2];
\end{tikzpicture}  \\[8mm]  
    
    E2  & \begin{tikzpicture}[baseline=(current bounding box.center)]
   \draw[fill=blue!100,draw=blue] (-1.2,0) circle [radius=0.18];
  \draw[fill=red!100,draw=red] (-0.9,-0.18) rectangle ++(0.36,0.36);
  \draw[fill=red!100,draw=red] (-0.2, 0) circle [radius=0.18];
\end{tikzpicture} & \begin{tikzpicture}[baseline=(current bounding box.center)]
 \draw[fill=blue!100,draw=blue] (-0.7,-0.18) rectangle ++(0.36,0.36);
  \draw[fill=green!100,draw=green] (-0.2,-0.18) rectangle ++(0.36,0.36);
  \draw[fill=green!100,green=blue] (-0.6,-0.5) circle [radius=0.18];
  \draw[fill=blue!100,draw=blue] (0,-0.5) ellipse [x radius=0.29, y radius=0.2];
  \draw[fill=pink!100,draw=pink] (-0.3,-0.9) ellipse [x radius=0.29, y radius=0.2];
\end{tikzpicture}  \\[3mm]  

    \bottomrule
     \label{tab-setting1}
    \end{tabular}
  \end{subtable}
  \hspace*{2.5em}
  \begin{subtable}[t]{0.2\textwidth}
    \centering
    \caption{Continual learning}
    \begin{tabular}{ccc}
      \toprule
      Index & Task 0 &  Task 1     \\
      \midrule
    E3  & \begin{tikzpicture}[baseline=(current bounding box.center)]
  \draw[fill=blue!100,draw=blue] (-1.2,0) circle [radius=0.18];
  \draw[fill=red!100,draw=red] (-0.85,-0.18) rectangle ++(0.36,0.36);
\end{tikzpicture} & \begin{tikzpicture}[baseline=(current bounding box.center)]
  \draw[fill=blue!100,draw=blue] (-0.7,-0.18) rectangle ++(0.36,0.36);
  \draw[fill=green!100,draw=green] (-0.2,-0.18) rectangle ++(0.36,0.36);
\end{tikzpicture}  \\[3mm]  

    E4  & \begin{tikzpicture}[baseline=(current bounding box.center)]
  \draw[fill=blue!100,draw=blue] (-1.2,0) circle [radius=0.18];
  \draw[fill=red!100,draw=red] (-0.9,-0.18) rectangle ++(0.36,0.36);
  \draw[fill=red!100,draw=red] (-0.2, 0) circle [radius=0.18];
\end{tikzpicture} & \begin{tikzpicture}[baseline=(current bounding box.center)]
  \draw[fill=blue!100,draw=blue] (-0.7,-0.18) rectangle ++(0.36,0.36);
  \draw[fill=green!100,draw=green] (-0.2,-0.18) rectangle ++(0.36,0.36);
\end{tikzpicture}  \\[3mm]  

   E5  & \begin{tikzpicture}[baseline=(current bounding box.center)]
  \draw[fill=blue!100,draw=blue] (-1.2,0) circle [radius=0.18];
  \draw[fill=red!100,draw=red] (-0.85,-0.18) rectangle ++(0.36,0.36);
\end{tikzpicture} & \begin{tikzpicture}[baseline=(current bounding box.center)]
  \draw[fill=green!100,green=blue] (-0.6,0) circle [radius=0.18];
  \draw[fill=green!100,draw=green](-0.2,-0.18) rectangle ++(0.36,0.36);
\end{tikzpicture}  \\[3mm]  

   E6  & \begin{tikzpicture}[baseline=(current bounding box.center)]
  \draw[fill=blue!100,draw=blue] (-1.2,0) circle [radius=0.18];
  \draw[fill=red!100,draw=red] (-0.9,-0.18) rectangle ++(0.36,0.36);
  \draw[fill=red!100,draw=red] (-0.2, 0) circle [radius=0.18];
\end{tikzpicture} & \begin{tikzpicture}[baseline=(current bounding box.center)]
  \draw[fill=green!100,green=blue] (-0.6,0) circle [radius=0.18];
  \draw[fill=green!100,draw=green](-0.2,-0.18) rectangle ++(0.36,0.36);
\end{tikzpicture}  \\[3mm] 

  E7  & \begin{tikzpicture}[baseline=(current bounding box.center)]
  \draw[fill=blue!100,draw=blue] (-1.2,0) circle [radius=0.18];
  \draw[fill=red!100,draw=red] (-0.85,-0.18) rectangle ++(0.36,0.36);
\end{tikzpicture} & \begin{tikzpicture}[baseline=(current bounding box.center)]
  \draw[fill=blue!100,draw=blue] (-0.7,-0.18) rectangle ++(0.36,0.36);
  \draw[fill=blue!100,draw=blue] (0,0) ellipse [x radius=0.29, y radius=0.2];
\end{tikzpicture}  \\[3mm]  

  E8  & \begin{tikzpicture}[baseline=(current bounding box.center)]
  \draw[fill=blue!100,draw=blue] (-1.2,0) circle [radius=0.18];
  \draw[fill=red!100,draw=red] (-0.9,-0.18) rectangle ++(0.36,0.36);
  \draw[fill=red!100,draw=red] (-0.2, 0) circle [radius=0.18];
\end{tikzpicture} & \begin{tikzpicture}[baseline=(current bounding box.center)]
  \draw[fill=blue!100,draw=blue] (-0.7,-0.18) rectangle ++(0.36,0.36);
  \draw[fill=blue!100,draw=blue] (0,0) ellipse [x radius=0.29, y radius=0.2];
\end{tikzpicture}  \\[3mm] 

 E9  & \begin{tikzpicture}[baseline=(current bounding box.center)]
  \draw[fill=blue!100,draw=blue] (-1.2,0) circle [radius=0.18];
  \draw[fill=red!100,draw=red] (-0.85,-0.18) rectangle ++(0.36,0.36);
\end{tikzpicture} & \begin{tikzpicture}[baseline=(current bounding box.center)]
  \draw[fill=blue!100,draw=blue] (-0.7,-0.18) rectangle ++(0.36,0.36);
  \draw[fill=pink!100,draw=pink] (0,0) ellipse [x radius=0.29, y radius=0.2];
\end{tikzpicture}  \\[3mm]  

  E10  & \begin{tikzpicture}[baseline=(current bounding box.center)]
  \draw[fill=blue!100,draw=blue] (-1.2,0) circle [radius=0.18];
  \draw[fill=red!100,draw=red] (-0.9,-0.18) rectangle ++(0.36,0.36);
  \draw[fill=red!100,draw=red] (-0.2, 0) circle [radius=0.18];
\end{tikzpicture} & \begin{tikzpicture}[baseline=(current bounding box.center)]
  \draw[fill=blue!100,draw=blue] (-0.7,-0.18) rectangle ++(0.36,0.36);
  \draw[fill=pink!100,draw=pink] (0,0) ellipse [x radius=0.29, y radius=0.2];
\end{tikzpicture}  \\[3mm] 

    \bottomrule
    \label{tab-setting2}
    \end{tabular}
  \end{subtable}
  \caption{List of the class settings in each experiment in this study. The orders of the shapes in the table align with those used in training (Better viewed in color).}
\end{table}

We compute the \emph{Mahalanobis} distances, $M_{c}$, that often applied for OSR \cite{liu2020few,lee2018simple}, between the testing sample features, including both close-sets and open-sets, with their closest class centers in the training data to measure their similarities, which can indicate the OSR performances and provide a better view of the feature learning.
As demonstrated in Equation \eqref{equ-mahalanobis}, $\mathbf{\mu}_{c}$ refers to the center of class $c$, which is the mean of all the features in $c$. $\Sigma_{c}$ is the covariance matrix that can infer the geometrical shape of the class cluster. 
In the following text, the sets of such distances for close-sets and open-sets are denoted using $\mathbf{M}_{in}$ and $\mathbf{M}_{out}$, respectively. 

\begin{equation}
    \begin{gathered}
     M_{c} = (\mathbf{z}_i - \mathbf{\mu}_{c})^T{\mathbf{\Sigma_{c}}}^{-1}(\mathbf{z}_i - \mathbf{\mu}_{c})    \\
     \mathbf{\mu}_{c} = \frac{1}{|\mathbf{Z}_c|}\sum_{\mathbf{z}_c \in \mathbf{Z}_c} \mathbf{z}_c, 
     \Sigma_{c} = (\mathbf{Z}_c-\mathbf{\mu}_{c})^T(\mathbf{Z}_c-\mathbf{\mu}_{c})
    \end{gathered}
     \label{equ-mahalanobis}
\end{equation}

To investigate the separateness of the inlier and outlier features, we compute the histograms on $\mathbf{M}_{in}$ and $\mathbf{M}_{out}$ and adopt the normalized distance measure in \cite{cha2002measuring} to quantify the overlaps between the two histograms. As demonstrated in Equation \eqref{equ-measure-hist}. $N_{bins}$ denotes the number of bins of both histograms, and $h_i$ and $g_i$ are the number of samples falling into each bin of $\mathbf{M}_{in}$ and $\mathbf{M}_{out}$, respectively. 
$|\mathbf{M}_{in}|$ and $|\mathbf{M}_{out}|$ represent the number of bins in each histogram.
It can be easily inferred that the larger $D_{hist}$ is, the more separated the two histograms are. 
$D_{hist}=1$ indicates that the two histograms have no overlap.

\begin{equation}
    D_{hist} = \sum_{i=1}^{N_{bins}} |\frac{h_i}{|\mathbf{M}_{in}|} - \frac{g_i}{|\mathbf{M}_{out}|}|
    \label{equ-measure-hist}
\end{equation}

\begin{figure}
    \centering
    \includegraphics[width=0.99\linewidth]{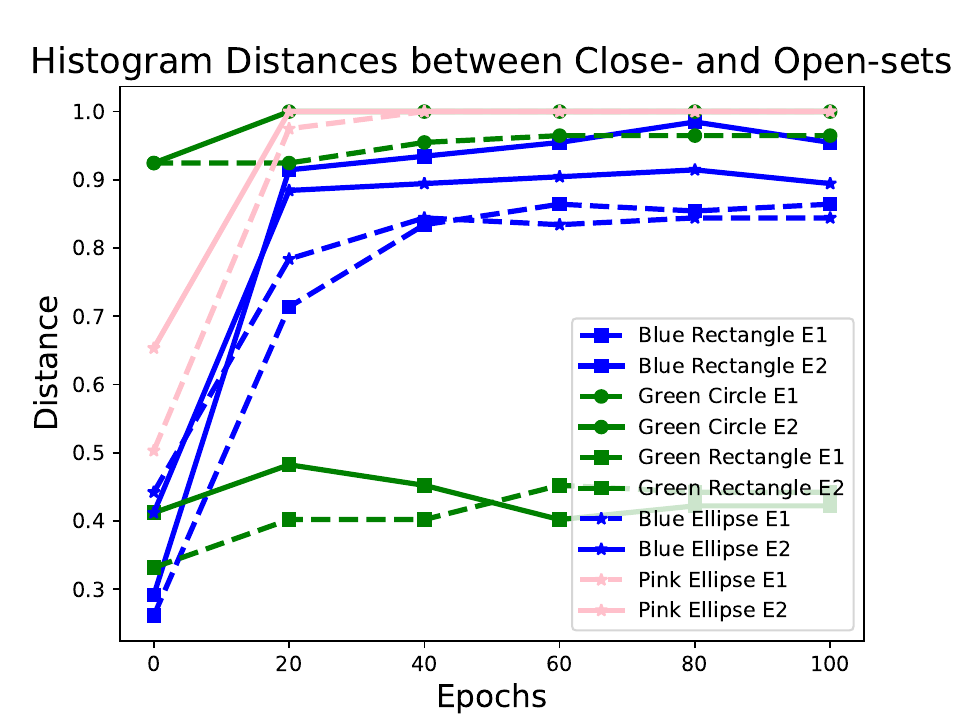}
    \caption{Plots of $D_{hist}$ for all open-sets. The colors and markers of the curves are in line with the colors and shapes of the data with exception of stars for ellipses. Plots for E1 are in dashed lines, while these for E2 are solid lines.}
    \label{fig-his-dis}
\end{figure}

As listed in Table \ref{tab-setting1}, we take blue rectangles, green rectangles, green circles, green ellipses, and pink ellipses as open-sets for E1 and E2. The same data will be used for learning new tasks for continual learning experiments in the following text. We compute $D_{hist}$ for all open-set classes with respect to their closest close-set classes. We plot the change of $D_{hist}$ with the training progress for all open-set classes in E1 and E2 in Figure \ref{fig-his-dis}.

It can be read from the plots that the $D_{hist}$ E2 are larger or at least equal to those in E1, indicating better OSR performances. It can lead to the conclusion that Learning diverse features can improve OSR. Additionally, for blue circles, we have observed during the experiments that they are closer to blue circles than red rectangles in both E1 and E2. We think the reason lies in that the colors can pose larger weights in the feature vectors than the shapes. The pink ellipses show high similarity with the red circles in E2. It is not hard to find the reasons that pink and red are close in the RGB channels, and ellipse and circle are similar. It can also explain the results of E10 in \ref{subsec-forward-transfer}.

\subsection{Continual Learning}

Following the above settings, we take the models in E1 and E2 as base models to learn new classes. 
For this, eight new continual learning models (E3-E10), whose class settings are listed in Table \ref{tab-setting2}, are trained. 
In the following text, the base tasks in E1 or E2 are referred as Task 0, and the following new task is Task 1. 
There are four groups of new task settings, which differ in whether the features in Task 0 are repeated and could be reused in the new task. Blue rectangle and green rectangle share the color features with Task 0, whereas green circles and green rectangles are the opposite and can be classified only through the shapes. The ellipse is a new shape from Task 0. E7 and E8 should learn this new shape for classification, whereas for E9 and E10, the red color feature could probably be reused.
To precisely record forgetting prevention and forward transfer learning abilities of the learned features in the base models, no extra continual learning approaches have been applied in all experiments here. Neurons are added on the output layers to fit the new number of classes.
Since so, the testing accuracy of the new models on the observed data in all experiments is around $0.5$, i.e., almost a guess, we therefore compare feature forgetting instead, which is closer to the topic of this study as well.


\subsubsection{Feature Forgetting}

\begin{table}[]
    \centering
    \small
    \begin{tabular}{cccc}
    \toprule
                & E1     &  E2 (2)  &  E2 (3)  \\
   \hline
    E3 / E4   & 0.0213  & 0.0216 &  0.0219 \\
    E5 / E6   & 0.0127  & 0.016  &  0.018  \\
    E7 / E8   & 0.017   & 0.0159 &  0.074  \\
    E9 / E10  & 0.0089  & 0.0169 &  0.0188 \\
    \bottomrule
    \end{tabular}
    \caption{Feature similarity for the data in Task 0 between the models before and after Task 1. The first row lists the base models, and E2 (2) and E2 (3) refer to the number of classes used for comparison, which are the two that overlap with E1 and all three classes in E2.}
    \label{tab-cka}
\end{table}

We compute the centered kernel alignment (CKA) \cite{kornblith2019similarity} between the features in Task 0 and Task 1 to quantify the feature forgetting. Higher CKA indicates higher similarity, which means less feature forgetting.
Since our dataset is very easy and the backgrounds are all black, the features can always pose very high CKA values, even though they are from different models or classes. We therefore change the background of the data used for feature reading to white. The results are shown in Table \ref{tab-cka}. 
When comparing E3 \& E4 with E5 \& E6, more features are forgotten in the later. In E3 \& E4, the color features can be reused, whereas in E5 \& E6, the models can rely only on the shapes. The colors, which can show higher weights in the feature vectors, can therefore be forgotten more in E5 \& E6. And the overall CKA values are lower.
Additionally, we believe the same reasons apply for comparing E7 \& E8 with E9 \& E10.

\subsubsection{Forward Transfer} \label{subsec-forward-transfer}

To evaluate the forward transfer ability of the models in E1 and E2, we use linear probing on the samples from Task 1 to learn a classifier and then evaluate the accuracy \cite{chen2023forgetting}. For this, all layers in the models except the newly initialized output layer are frozen.
The results are in Table \ref{tab-linear-probe}.

\begin{table}[]
    \centering
    \small
    \begin{tabular}{ccccc}
    \toprule
           & E3/E4 &  E5/E6 & E7/E8  &  E9/E10    \\
     \hline
    E1    &  $90.83$  &  $70.5$ &  $68.67$  &  $53.17$  \\
    E2    &  $92.67$  &  $87.33$ & $68.67$  &  $95.5$  \\
    \bottomrule
    \end{tabular}
    \caption{Linear probing accuracy (in \%) for Task 1 data in E3, E4, E5, E6, E7, and E8 for evaluating the forward transfer ability of the base models in E1 and E2. The base models are noted in the first column.}
    \label{tab-linear-probe}
\end{table}

For E3 / E4, since the testing data can be classified with only colors, and the blue color is learned in both base models, the results therefore don't show a very significant difference. 
For E5/ E6, it is completely different. While the colors are no longer useful for the classification, the accuracy in E6 is much higher than in E5 because the base model E2 has better learned the shapes. However, it is not completely guessing in E5. It can indicate that E1 has encoded shape features, which can be however worse than E2. 
For E7 and E8, the numbers are the same. For new shapes that do not exist in the training data, the ability to classify new data can be independent on the base models. 
Lastly, for E9 and E10, since the ellipses are pink, which are close to the red circles in E2. E2 models can then be well transferred to the new data, which has been mentioned in \ref{subsec-osr} above. However, altogether with E7 and E8, where the blue ellipses cannot show such positive effects. We hence think the reason is that the features are not disentangled in the base models.

In summary, learning more features from the observed data is helpful for learning new data in continual learning when there are overlaps between the features in the observed and new data. However, in practice, the new tasks are always unknown, which means it cannot be foreseen which features are useful for the future tasks. It is therefore reasonable to learn as diverse features as possible from the observed data while maintaining the performance of the tasks so far. Besides, feature disentanglement can be important for learning new tasks because the observed data can then be better compared with the new data, and overlapped features can be applied to the new tasks. We leave this for future work.

%% file: sec/7_conclusion.tex
\section{Conclusion \& Future Work} \label{sec-conclusion}

In this study, we have investigated the effects of learning diverse features in supervised classifiers for open set recognition and continual learning. The results can demonstrate that more diverse features can be helpful for detecting open-sets, reducing feature forgetting for past tasks in continual learning, and learning new tasks. 

However, our experiments are based on synthetic datasets, which are of small volume. In future work, we hope new experiments can be designed for real and complex data. Furthermore, feature disentanglement remains an interesting topic to explore its connection with OSR and continual learning. We hypothesize that disentangling features can be positive for both OSR and continual learning.